\documentclass[11pt,a4paper]{scrartcl}

\usepackage{graphicx}
\usepackage{subfigure}
\usepackage{amsfonts}
\usepackage{amsthm}
\usepackage{amssymb}
\usepackage{amstext}
\usepackage{amsmath}
\usepackage{amsthm}
\usepackage{pstricks}
\usepackage{pspicture}
\usepackage{amscd}
\usepackage{setspace}
\usepackage{bigstrut}

 \newtheorem{proposition}{Proposition}

	\newtheorem{corollary}{Corollary}
	\newtheorem{axiom}{Axiom}

\begin{document}
\title{Recent advances on inconsistency indices for pairwise comparisons --- a commentary\footnote{This manuscript has been published as: Brunelli M., Recent advances on inconsistency indices for pairwise comparisons --- a commentary, \emph{Fundamenta Informaticae}, 144(3--4), 321--332, 2016. doi:10.3233/FI-2016-1338.}}
\author{
{Matteo Brunelli}
\\
{\normalsize  SAL, Department of Mathematics and Systems Analysis} \\
{\normalsize Aalto University}, {\normalsize Otakaari 1,
Aalto FI-00076, Finland}
\\ {\normalsize e-mail:
\texttt{matteo.brunelli@aalto.fi}}
\vspace{0.3cm}\\
%
%{\bf Andrew Critch}
%\\
%{\normalsize Department of Mathematics} \\
%{\normalsize University of California}, {\normalsize Berkeley, CA
%94720, United States}
%\\ {\normalsize e--mail:
%\texttt{critch@math.berkeley.edu}}
%\vspace{0.3cm}\\
%
%{\bf Michele Fedrizzi}
%\\
%{\normalsize  Department of Computer and Management Sciences} \\
%{\normalsize University of Trento}, {\normalsize Via Inama 5, I-38122 Trento, Italy}
%\\ {\normalsize e--mail:
%\texttt{michele.fedrizzi@unitn.it}}
}
\date{}

\maketitle \thispagestyle{empty}
% ---------------------------------------------------------------------

\begin{center}
{\bf Abstract }
\end{center}

{\small \noindent  This paper recalls the definition of consistency for pairwise comparison matrices and briefly presents the concept of inconsistency index in connection to other aspects of the theory of pairwise comparisons. By commenting on a recent contribution by Koczkodaj and Szwarc, it will be shown that the discussion on inconsistency indices is far from being over, and the ground is still fertile for debates.}

 \vspace{0.3cm}
 \vspace{0.3cm}

% ---------------------------------------------------------------------

\section{Introduction}

In problems of choice, decision makers often face the selection of a single best alternative from a feasible set. In modern methodologies for decision making, as for instance the Analytic Hierarchy Process (AHP) \cite{Saaty1977}, the use of pairwise comparisons between alternatives has been strongly advocated. Comparing two alternatives at a time and expressing the degree of preference of one to the other, is supposed to simplify the original problem by decomposing it into smaller and more easily tractable ones. 

The most widely known representation of valued preferences is based on the pairwise comparison matrices used by Saaty in the AHP. Given a finite non-empty set of alternatives, $\{ \theta_{1},\ldots,\theta_{n} \}$, a \emph{pairwise comparison matrix} is a matrix $\mathbf{A}=(a_{ij})_{n \times n}$, with $a_{ij}a_{ji}=1 \; \forall i,j$, where $a_{ij}>0$ represents the degree of preference of $\theta_{i}$ to $\theta_{j}$. Moreover, a pairwise comparison matrix is called \emph{consistent} if the following condition holds,
\begin{equation}
\label{eq:cons}
a_{ik}=a_{ij}a_{jk}~~~\forall i,j,k.
\end{equation}
Such a condition is the formalization of a desirable property of cardinal transitivity. Namely, the value of each direct comparison $a_{ik}$ is precisely backed up by all indirect comparisons $a_{ij}a_{jk}$. For example, if $\theta_{i}$ is considered twice as good as $\theta_{j}$ which, in turn, is also considered twice as good as $\theta_{k}$, it appears reasonable that $\theta_{i}$ be four times as good as $\theta_{k}$. Furthermore, condition (\ref{eq:cons}) can be restated in the following equivalent ways:
\begin{itemize}
	\item There exists a (priority) vector $\mathbf{w}=(w_{1},\ldots,w_{n})$ such that $a_{ij}=w_{i}/w_{j}~\forall i,j$
	\item $\mathbf{A}$ has unitary rank, i.e. $\text{rank}(\mathbf{A})=1$
	\item The Perron-Frobenius eigenvalue of $\mathbf{A}$ is equal to $n$, i.e. $\lambda_{\max}=n$.
\end{itemize}
The scientific debate has centered on the methods used to quantify the deviation between the preferences of a decision maker and the condition of full consistency. Such quantification is usually done by functions which, in the literature, are called {inconsistency indices}.

Mathematically speaking, an \emph{inconsistency index} is a function $I$ which maps pairwise comparison matrices into the real line, so that $I(\mathbf{A})\in \mathbb{R}$ is an estimation of the inconsistency of $\mathbf{A}$. Usually, the greater $I(\mathbf{A})$, the higher the level of inconsistency. For some inconsistency indices, threshold values $\tau$ have been proposed, so that, if $I(\mathbf{A}) > \tau$, then the matrix $\mathbf{A}$ is considered too inconsistent and the decision maker has to revise his judgments. It is worth commenting that there is not a meeting of minds on what the thresholds should be and how they could be estimated. For example, Saaty proposed $\tau=0.1$ for his index $CR$, where 0.1 means that all the matrices with inconsistency smaller than the 10\% of the inconsistency of a random matrix should be accepted. Koczkodaj, instead, proposed $\tau=1/3$ for the index $K$. Interestingly, many indices have been introduced in the literature without studying their thresholds.

\begin{figure}[h!tbp]
	\centering
		\includegraphics[width=0.50\textwidth]{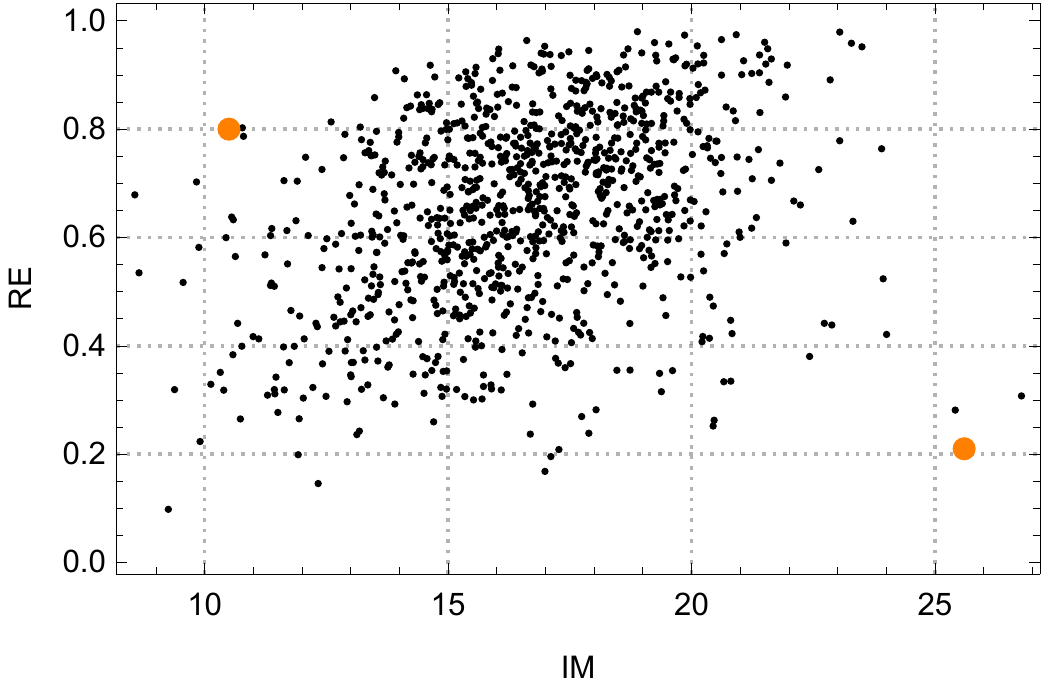}
	\caption{Indices $IM$ \cite{GassRapcsak2004} and $RE$ \cite{Barzilai1998} computed on 2,000 pairwise comparison matrices of order 6.}
	\label{fig:a_b}
\end{figure}

A large number of indices have been proposed; ten of them were numerically analyzed \cite{BrunelliCanalFedrizzi}, and some more were proposed more recently \cite{Kulakowski2015}. Some indices were even proven proportional \cite{BrunelliCritchFedrizzi2013} or equivalent \cite{Kulakowski2014} to each other, but what is more remarkable is how differently they may behave. Figure \ref{fig:a_b} shows an example where two inconsistency indices are compared via numerical simulations: each point in the scatter-plot corresponds to a pairwise comparison matrix and its coordinates on the axes represent its inconsistency evaluated by the two inconsistency indices proposed by Barzilai \cite{Barzilai1998} and Gass and Rapcs\'{a}k \cite{GassRapcsak2004}, respectively.
Two significant cases are highlighted with a larger and clearer dot. In both cases an index evaluates a matrix as extremely inconsistent, whereas the other one indicates that the same matrix is almost consistent. Such discordant behavior is not limited to the two indices presented in the example. As different indices often yield different results and conclusions, choosing among them really makes a difference. All this just adds up to the need of a deeper understanding of inconsistency indices.

In the next section, the paper will first recall the central role of inconsistency indices in the theory of pairwise comparison matrices. Thereafter, it will discuss the recent contribution by Koczkodaj and Szwarc \cite{KoczkodajSzwarc2014}, which presented a critical assessment of a number of inconsistency indices implying their unsoundness as inconsistency estimators. By challenging their conclusions, this paper will hopefully put forward a more tolerant viewpoint.

\section{The central role of inconsistency indices}

The role played by inconsistency indices has not been limited to the mere estimation of the irrationality of pairwise comparisons. With its ramifications, the concept of consistency, and of its counterpart inconsistency, has permeated the whole theory of pairwise comparisons.

A fundamental step in the exploitation of the information contained in a pairwise comparison matrix is the synthesis of this latter into a \emph{priority vector} $\mathbf{w}=(w_{1},\ldots,w_{n})$. Many of the methods used to estimate $\mathbf{w}$ are mathematically connected to an index of inconsistency. For example, the eigenvector method is associated with $\lambda_{\max}$ and consequently to $CI$ \cite{Saaty1977}. Another example is the Geometric Consistency Index \cite{AguaronMoreno2003}, which is connected to the so-called Geometric Mean method used to determine $\mathbf{w}$ \cite{CrawfordWilliams1985}. In addition, many methods to estimate $\mathbf{w}$ seek for the consistent matrix $( w^{*}_{i}/w^{*}_{j} )_{n \times n}$ which is the closest to $\mathbf{A}$. In the literature, the distance between $( w^{*}_{i}/w^{*}_{j} )_{n \times n}$ and $\mathbf{A}$ has often been interpreted as an estimation of the inconsistency of $\mathbf{A}$.

When the number of alternatives grows, it can happen that a decision maker is unable to express all the $n(n-1)/2$ independent pairwise comparisons necessary to complete the matrix.  Many methods are concerned with the optimal completion of matrices. That is, given a subset of comparisons, how can we estimate the missing ones?
One very common approach is that of minimizing an inconsistency index. In this case the missing values are the variables. For example, given the following incomplete pairwise comparison matrix,
\begin{equation}
\mathbf{A}=
\begin{pmatrix}
1   & 2 & a_{13}   & a_{14} \\
1/2 & 1 & 1/3 & 1 \\
1/{a_{13}} & 3 & 1   & 2 \\
1/{a_{14}} & 1 & 1/2 & 1
\end{pmatrix},
\end{equation}
one might want to solve the optimization problem
\begin{equation}
\label{eq:opt}
\begin{aligned}
& {\text{minimize}}
& & I(\mathbf{A}) \\
& \text{subject to}
& & a_{13}, a_{14} >0 \,  
\end{aligned}
\end{equation}
to find the most suitable values of $a_{13}$ and $a_{14}$ with respect to a given inconsistency index $I$.
\noindent Bozoki et al. \cite{BozokiEtAl2010}, Shiraishi et al. \cite{ShiraishiEtAl1999}, Koczkodaj \cite{Koczkodaj1999} and Lamata and Pelaez \cite{LamataPelaez2002} all proposed optimization problems in the form of (\ref{eq:opt}). The importance of incomplete preferences has been of ever increasing interest in the literature, as also supported by the recent surveys by Ishizaka and Labib \cite{IshizakaLabib2011} and Ure\~{n}a et al. \cite{UrenaEtAl2015}.

Optimization problems have also been used recently to improve the consistency of preferences when these latter ones are excessively inconsistent. Examples are the optimization problems proposed by Dong and Herrera-Viedma \cite{DongHerrera-Viedma2015} and Dong et al. \cite{DongEtAl2015} to mitigate the inconsistency which arises, for example, due to the due to the nature of Saaty's scale.

There exists \emph{alternative representations} of valued preferences where the pairwise comparisons are equivalently expressed on other representation domains. The foremost are (i) reciprocal relations \cite{DeBaetsEtAl2006}, often called fuzzy preference relations in the fuzzy sets literature \cite{Tanino1984}, and (ii) additive preference relations, loosely related with Fishburn's skew-symmetric additive representation of preferences \cite{Fishburn1999}. For these representations of preferences, inconsistency indices have also been proposed. One can consider, for instance, those introduced by Barzilai \cite{Barzilai1998}, Ji and Jiang \cite{JiJiang2003} and Yuen \cite{Yuen2012} for additive preferences. Given the existence of transformations between different representations \cite{CavalloDApuzzo2009}, results obtained in one context might be directly applicable in another.

Inconsistency indices have also been considered in conjunction with group decisions and consensus reaching, e.g. by Chiclana et al. \cite{ChiclanaEtAl2008} and Dong et al. \cite{DongEtAl2010}. This seems to support the thesis that aspects of pairwise comparisons which were studied separately are, conversely, intertwined.

Especially in light of the most recent development, it should be clear that inconsistency indices play a central role in the theory of valued preference relations. It naturally follows that a fair and rigorous analysis is just necessary.

\section{(Re)opening the debate on inconsistency indices}

In spite of the variety of indices proposed in the literature, until recent times there were no axiomatic studies.
%The idea behind an axiomatic study would be that of defining a minimum set of necessary properties for a function $I$ to be considered an inconsistency index.
Two recent proposals of axiomatization were by Brunelli and Fedrizzi \cite{BrunelliFedrizziAxioms} \footnote{Preliminary results on this set of axioms were already presented as ``M. Brunelli and M. Fedrizzi, Characterizing properties for inconsistency indices in the analytic hierarchy process, \emph{Abstracts of MCDM} 2011", and ``M. Brunelli and M. Fedrizzi, Characterizing properties for inconsistency indices in the AHP, \emph{Proceedings of ISAHP 2011}"} and by Koczkodaj and Szwarc \cite{KoczkodajSzwarc2014}. Besides presenting separate axiomatic systems, the two proposals differ in the conclusions, with Koczkodaj and Szwarc being more drastic. For sake of brevity, the reader's familiarity with Koczkodaj and Szwarc's results is assumed, even though the main concepts will be recalled.

In their paper, Koczkodaj and Szwarc proposed a set of three axioms. The first two axioms are simple regularity conditions while the third seems to be more constraining. These three axioms are in the form of reasonable properties and serve to implicitly define inconsistency indices. In line with this, the authors claimed that their axiomatic system ``allows us to define proper inconsistency indicators". Therefore, one might suppose that if a function satisfies their axioms, then they should grant it the status of ``proper" inconsistency index.

\subsection{The triad and the matrix}

The interpretation of the third axiom of Koczkodaj and Szwarc is problematic, because of its ambiguous definition and the soundness of the adduced motivation. According to them \cite{KoczkodajSzwarc2014}:
\begin{quote}
``It is a reasonable expectation that the worsening of a triad, used in the definition of consistency, cannot make the entire matrix more consistent".
\end{quote}
The following example challenges their expectation. Consider the matrix

\[
\begin{pmatrix}
1    & 2 & 2 & 4 & 7 \\
1/2  & 1 & 4 & 1 & 3 \\
1/2  & 1/4 & 1 & 1 & 4\\
1/4  & 1 & 1 & 1 & 2\\
1/7  & 1/3 & 1/4 & 1/2 & 1 \\
\end{pmatrix}.
\]
The entry $a_{15}$ appears in the triads $(a_{12},a_{25},a_{15})$, $(a_{13},a_{35},a_{15})$ and $(a_{14},a_{45},a_{15})$. Let us write them down more explicitly.
\begin{align*}
(a_{12},a_{25},a_{15}) = (2 , 3 , 7)  \\
(a_{13},a_{35},a_{15}) = (2 , 4 , 7)  \\
(a_{14},a_{45},a_{15}) = (4 , 2  ,7)
\end{align*}
Now, if we slightly increase the value of $a_{15}$ from, say, $7$ to $7.5$, this worsens the local inconsistency of the triad $(a_{12},a_{25},a_{15})$. By following the assumption by Koczkodaj and Szwarc, this should \emph{not} imply an enhancement of the consistency of the entire matrix. It occurs that this assumption neglects the incidence of $a_{15}$ on the inconsistency of other triads. In fact, in this case, both $(a_{13},a_{35},a_{15})$ and $(a_{14},a_{45},a_{15})$ benefit from such a change and become less inconsistent.
The example with a matrix of order $5$ was purely illustrative, and drawing from basic graph theory we know the following proposition.
\begin{proposition}
Given a pairwise comparison matrix $\mathbf{A}$ of order $n$ and its triples $(a_{ij},a_{jk},a_{ik})~\forall 1 \leq i<j<k \leq n$, then a single comparison $a_{ij}~(i \neq j)$ appears in exactly $(n-2)$ triples.
\end{proposition}
\noindent Furthermore, one could easily build up an example of matrix of order $n$ where, by changing one entry and its reciprocal we worsen one single triad but we ameliorate all other $(n-3)$ of them. Hence, in light of the fact that, except for matrices of order $3$, changing one entry has effects on a plurality of triads, the assumption proposed by Koczkodaj and Szwarc becomes unattainable and therefore useless.

\subsection{The role of an axiomatic system}
\begin{figure}[htbp]
	\centering
		\includegraphics[width=0.45\textwidth]{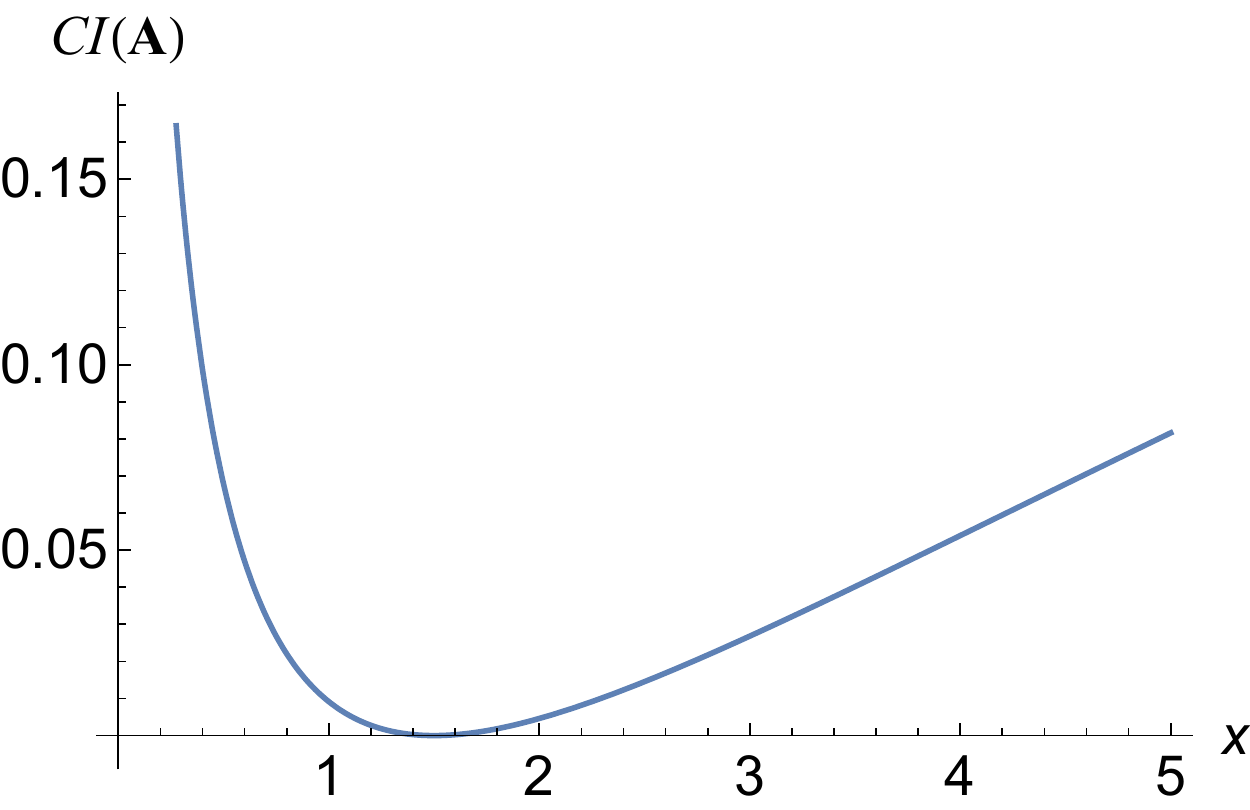}
	\caption{Quasi-convexity of $CI(\mathbf{A})$ with respect to $x$.}
	\label{fig:quasi-convexity-CI}
\end{figure}

As correctly claimed by Koczkodaj and Szwarc \cite{KoczkodajSzwarc2014}, in this context, an axiomatic system serves to define a set of functions which are, in principle, all suitable to represent a given characteristic of a mathematical object. Many other axiomatic systems work in a similar way as, for instances, the axiomatic definition of distance in metric spaces or of norm in linear spaces. It should follow that, if a function $I$ satisfies their three axioms, then this same function ought to be respected as an inconsistency index (at the very least by the authors who proposed the axioms).\\
In spite of the ambiguous formulation (especially concerning the third axiom), by considering a $3 \times 3$ pairwise comparison matrix
\[
\mathbf{A}_{3 \times 3}
=
\begin{pmatrix}
1         & a_{ij} & a_{ik} \\
1/a_{ij}  & 1      & a_{jk} \\
1/a_{ik}  & 1/a_{jk}& 1 
\end{pmatrix},
\]
the axiomatic system by Koczkodaj and Szwarc can be formalized as follows.
\begin{axiom}
If $\mathbf{A}_{3 \times 3}$ is consistent, then $I(\mathbf{A}_{3 \times 3})=0$.
\end{axiom}
\begin{axiom}
$I(\mathbf{A}_{3 \times 3}) \in [0,1[$.
\end{axiom}
\begin{axiom}
$I(\mathbf{A}_{3 \times 3})$ is a quasi-convex function with respect to $a_{ij},a_{jk},a_{ik}>0$, with global minimum attained when $a_{ik}=a_{ij}a_{jk}$.
\end{axiom}
In their research, Koczkodaj and Szwarc concluded that the index
\[
CI(\mathbf{A})=\frac{\lambda_{\max}-n}{n-1},
\]
is inadequate to capture inconsistency, but they do not mention whether it satisfies the axioms or not.
However, by examining the literature one finds that the behavior of $CI$ with respect to the entries of $\mathbf{A}$ was already studied by others. Bozoki et al. \cite{BozokiEtAl2010} examined some properties of generalized convexity of $CI$ and referred to previous results by Kingman \cite{Kingman1961} and Aupetit and Genest \cite{AupetitGenest1993}. Considering a more relaxed version of the second axiom by Koczkodaj and Szwarc, the following is a direct corollary of the above mentioned studies.
\begin{corollary}
If we substitute Axiom 2 with the more relaxed requirement that $I(\mathbf{A}_{3 \times 3}) \in [0,\infty[$, then index $CI$ satisfies all the properties proposed by Koczkodaj and Szwarc.
\end{corollary}
\noindent For an illustrative example, consider the matrix
\[
\mathbf{A}=
\begin{pmatrix}
1   & 3 & x \\
1/3 & 1 & 1/2 \\
1/x & 2 & 1
\end{pmatrix}.
\]
Figure \ref{fig:quasi-convexity-CI} illustrates the quasi-convex behavior of $CI(\mathbf{A})$ as a function of $x$.
For the proofs that several other inconsistency indices satisfy the axiomatic framework by Koczkodaj and Szwark, the reader can refer to the axiomatic study by Brunelli and Fedrizzi \cite{BrunelliFedrizziAxioms}.\\
Moreover, one shall note that the index $CI$ could be normalized, e.g. by dividing its value by its upper bound as specified by Aupetit and Genest \cite{AupetitGenest1993}, and thus, for practical purposes, it would satisfy all the axioms by Koczkodaj and Szwarc. In conclusion, it appears that $CI$ and many other indices satisfy the axioms proposed by Koczkodaj and Szwarc and, by using their own words, ought to be considered proper inconsistency indices.

\subsection{One rotten apple does not necessarily spoil the barrel}
To support their criticisms against $CI$, and incidentally towards a number of other inconsistency indices, Koczkodaj and Szwarc considered the following pairwise comparison matrix, where the order $n$ is not fixed,
\begin{equation}
\label{eq:A_KS}
\mathbf{A}_{KS}(x)=
\begin{pmatrix}
1 & 1 & \cdots & 1 & x \\ 
1 & 1 & \cdots & 1 & 1 \\
\vdots & \vdots & \ddots & \vdots & \vdots \\ 
1 & 1 & \cdots & 1 & 1 \\
1/x & 1 & \cdots & 1 & 1
\end{pmatrix}.
\end{equation}
Their main result is that, it does not matter the value of $x>0$, when $n$ grows to infinity, $CI(\mathbf{A}_{KS}(x))$ tends to zero. The authors considered this a \emph{reductio ad absurdum} and used it to invalidate $CI$ and all indices sharing the same property. Seemingly, Koczkodaj and Szwarc must have thought that the absurdity was self evident, since they did not further motivate it.\\
Let us try to interpret the result in a semi-real context, just with the restrictions imposed by reasoning in a limiting sense. Especially when $n$ is large, it is very plausible that the judgment $a_{1n}=x$ is the outlier, since it is not in accordance with all the others. Hence, $a_{1n}=x$ can be seen as an error made by the decision maker in expressing opinions.
Let us now propose the short story of Laura and her awful grade. Its resemblance with the problem at stake will hopefully be clear to the reader.
\begin{quote}
Suppose that Laura is a student in a school where it is possible to get an arbitrarily bad grade. Further suppose that one day Laura gets a dreadful grade, say $x$.
%Even though the grade can be as dreadful as she could have imagined, it remains one single grade.
After this unfortunate episode, Laura keeps on getting magnificent grades in the exams on the same subject. Is it fair to allow her, at some point in the future (assume that she has infinitely much time), to be able to make up for the initial bad grade and eventually be considered a sufficiently good student?
\end{quote}
This is definitely a moot point. Contrasting opinions on the issue might exist, and the mere existence of different, yet respectable opinions, on a proposal seem to clash with the very same proposal being declared absurd.
One simple argument in favor of Laura is that, if a slightly bad grade can be made up for, then also worse grade can be made up for, just in a longer time. And if not, where would one draw the (arbitrary) line between repairable and unrepairable bad grades? Furthermore, it is noteworthy that a school using arithmetic mean---which is often the case---allows compensations among  grades and Laura would be allowed to repair the damage made by the initial bad grade.

Likewise, when applied to the case of inconsistency of preferences, the story of Laura shows us that the \textit{absurdum} is not so absurd, and that approaches contemplating the possibility of compensating a local inconsistency by means of many consistent comparisons ought to be respected.

Moreover, on a more computational note, the \emph{relative incidence} of a wrong judgment, or of a relatively small number of judgments, decreases with the growth of the order of the matrix. In fact, the relative importance of few outliers in processes such as the determination of the priority vector diminishes as the order of the matrix grows. Indeed, this type of argument is borrowed from statistical reasoning, where it is accepted that the incidence of noisy observations can be mitigated by the growth of the sample size. A pairwise comparison matrix is not a completely different problem since more comparisons are required than those necessary to elicit a weight vector. Besides, many methods for determining the weight vector are borrowed from statistics, as for instance the Least Squares and the Logarithmic Least Squares which, in a way, consider the entries of $\mathbf{A}$ as the sample from which $\mathbf{w}$ is estimated.

\subsection{On the index $K$}

Koczkodaj and Szwarc supported the use of the following index \cite{Koczkodaj1994,Koczkodaj1993}
\begin{equation}
\label{eq:K}
K(\mathbf{A})=
\max_{i < j < k} \left\{
\min \left\{ \bigg| 1-\frac{a_{ij}a_{jk}}{a_{ik}}\bigg| , \bigg| 1-\frac{a_{ik}}{a_{ij}a_{jk}} \bigg| \right\} \right\},
\end{equation}
on the ground that the alleged \textit{reductio ad adsurdum} does not apply to it. Hereafter, we shall see that even index $K$, in some particular cases, gives rise to doubts on its capacity to capture inconsistency.
This can be shown starting with the following example. Considering the matrix $\mathbf{A}_{KS}(x)$ in (\ref{eq:A_KS}) and fixing, for example, $x=2$ we obtain, 
\begin{equation}
\label{eq:A}
\mathbf{A}_{KS}(2)=\begin{pmatrix}
1 & 1 & \cdots & 1 & 2 \\ 
1 & 1 & \cdots & 1 & 1 \\
\vdots & \vdots & \ddots & \vdots & \vdots \\ 
1 & 1 & \cdots & 1 & 1 \\
1/2 & 1 & \cdots & 1 & 1
\end{pmatrix}.
\end{equation}
With the currently proposed acceptance threshold set at $K < 1/3$ \cite{KoczkodajSzwarc2014}, one computes $K(\mathbf{A}_{KS}(2))=1/2$ for all $n$, which means that $\mathbf{A}_{KS}(2)$ is \emph{not} sufficiently consistent and thus needs revision. Besides, the peculiar feature of $K$ is its insensitivity to $n$. In fact, it is trivial to prove that, in the case of $\mathbf{A}_{KS}(2)$, $K$ is invariant with respect to $n$. Seen from this point of view, and considering the pairwise comparison matrix in (\ref{eq:A}) it seems more realistic that the value of inconsistency of $\mathbf{A}_{KS}(2)$ varies with $n$.

Let us consider the following two matrices, to show that the superiority of $K$ ought not to be taken for granted.
\[
\mathbf{A}_{1}=\begin{pmatrix}
1 & 1 & 1 & 1 & 2.001 \\ 
1 & 1 & 1 & 1 & 1 \\
1 & 1 & 1 & 1 & 1 \\ 
1 & 1 & 1 & 1 & 1 \\
\frac{1}{2.001} & 1 & 1 & 1 & 1
\end{pmatrix}
\hspace{1cm}
\mathbf{A}_{2}=
\begin{pmatrix}
1 & 1 & 2 & 1 & 2 \\ 
1 & 1 & 1 & 2 & 1 \\
1/2 & 1 & 1 & 1 & 2 \\ 
1 & 1/2 & 1 & 1 & 1 \\
1/2 & 1 & 1/2 & 1 & 1
\end{pmatrix}
\]
Although $K(\mathbf{A}_{1}) > K(\mathbf{A}_{2})$, it would also be legitimate to expect an index $I$ to yield the opposite, $I(\mathbf{A}_{1}) < I(\mathbf{A}_{2})$. The situation becomes more evident if we consider all the possible triads $(a_{ij},a_{jk},a_{ik})$ with $i<j<k$. In the case of $\mathbf{A}_{1}$ only 3 triads out of 10 are inconsistent and all the others appear perfectly consistent. Conversely, in $\mathbf{A}_{2}$ \emph{all} the triads are inconsistent.\\
The findings are even more evident if we reckon that this example can be generalized with respect to the order of the matrix and the values of the non-unitary entries.
\[
\mathbf{A}_{3}=\begin{pmatrix}
1 & 1 & \cdots & 1 & \alpha + \epsilon \\ 
1 & 1 & \cdots & 1 & 1 \\
\vdots & \vdots & \ddots & \vdots & \vdots \\ 
1 & 1 & \cdots & 1 & 1 \\
\frac{1}{\alpha + \epsilon} & 1 & \cdots & 1 & 1
\end{pmatrix}
\hspace{1cm}
\mathbf{A}_{4}=
\begin{pmatrix}
1 & 1 & \alpha & 1 & \alpha & \cdots \\ 
1 & 1 & 1 & \alpha & 1 & \cdots \\
1/\alpha & 1 & 1 & 1 &\alpha & \cdots \\ 
1 & 1/\alpha & 1 & 1 & 1 & \cdots \\ 
\vdots & \vdots & \vdots & \vdots & \ddots 
\end{pmatrix}
\]
For $\epsilon > 0$, it is $K(\mathbf{A}_{3})>K(\mathbf{A}_{4})$. Note moreover that, if we consider an arbitrarily small $\epsilon >0$, we have $\alpha \approx \alpha + \epsilon$, and the differences in the violations of the transitivities become negligible. Therefore, since the extents of violations are approximately equal, it makes more sense to count the number of violations instead of their intensities.
In $\mathbf{A}_{3}$ the number of triads containing inconsistencies is equal to $(n-2)$ while the total number of triads is $\binom{n}{3}$. Their ratio, simplified, is
\[
\frac{\text{number of inconsistent triads}}{\text{number of triads}}=\frac{n-2}{\binom{n}{3}}=\frac{6}{(n-1)n}.
\]
This ratio, and with it the relative importance of the inconsistent triads with respect to the whole, tends to zero as $n$ progresses to infinite. Nonetheless, by considering $K$ we have that the inconsistency of $\mathbf{A}_{3}$ is always greater than the inconsistency of $\mathbf{A}_{4}$ (where \emph{all} the triads are intransitive!).\\
By writing $\phi(a_{ij},a_{jk},a_{ik})= \min \left\{ |1-a_{ik}/(a_{ij}a_{jk})|, |1- (a_{ij}a_{jk})/a_{ik}| \right\}$ we can rewrite
\[
K(\mathbf{A})= \max_{i<j<k} \left\{ \phi(a_{ij},a_{jk},a_{ik}) \right\}
\]
and thanks to the properties of $\max$ we can enunciate the following simple proposition.
\begin{proposition}
Index $K$ is not a strictly monotone increasing function with respect to the $\phi(a_{ij},a_{jk},a_{ik})$, except for the greatest of them.
\end{proposition}
In words, the peculiar behavior of $K$, illustrated above, is due to the fact that index $K$ considers the most inconsistent triad only, whereas many other indices average the inconsistencies coming from different triads. 

\section{Conclusions}
Let us momentarily forget about the subject matter and instead consider the following two norms of the vector $\mathbf{x}=(x_{1},\ldots,x_{n}) \in \mathbb{R}^{n}$,
\[
N_{1}(\mathbf{x})=\frac{1}{n} \sum_{i=1}^{n}|x_{i}|,  \hspace{1cm} N_{2}(\mathbf{x})=\max_{i=1,\ldots,n}|x_{i}| .
\]
Ideally, as also proposed by Horn and Johnson \cite{HornJohnson1985}, a norm should capture the ``size'' of a vector. The two norms $N_{1}$ and $N_{2}$ capture two views on the ``size'' of $\mathbf{x}$; namely, $N_{1}$ averages the component-wise contributions whereas $N_{2}$ considers only the greatest one. It follows that they can give very different answers to the question concerning the ``size'' of the vector $\mathbf{x}$. Nonetheless, they are both norms as they respect the axiomatic conditions for norms.\\
Considering the vector $ \bar{\mathbf{x}} = (\delta,0,0,\ldots,0,0)  \in \mathbb{R}^{n}$, one can see that it does not matter how large $\delta > 0$ is, as we increase $n$, $N_{1}(\bar{\mathbf{x}})$ will always tend to zero. Still, nobody has ever used this argument to invalidate $N_{1}$, and both $N_{1}$ and $N_{2}$ coexist under the same umbrella. It has been up to applied mathematicians, economists, computer scientists and others, to distinguish between different norms, be aware of their properties, and select the one which seems the most suitable for a given context or situation.

It is straightforward to transpose this discussion to the realm of inconsistency indices. If some researchers propose a sound axiomatic system in which an inconsistency index fits, then this inconsistency index should not be considered incorrect, in the same way as neither $N_{1}$ nor $N_{2}$ should be classified as an incorrect norm. 

Different axiomatic systems for inconsistency indices have been proposed. They can be discussed and hopefully improved. Within these systems there exists a wealth and a diversity of indices which should be seen as a richness, and not as a threat. This paper showed examples where the behavior of index $K$ was questionable. Shall we therefore label it as an unsound inconsistency index? The answer of the author of this paper is negative. Different indices have different strong points and weaknesses, but each of them represents a valuable point of view on the same phenomenon: inconsistency of preferences.

{\small

\section*{Acknowledgements}

The author is grateful to the anonymous reviewers for their constructive comments.

}

%\bibliographystyle{abbrv}
%
%{ \footnotesize
%
%\bibliography{Bib_comments_consistency_rev} }

\end{document}